\documentclass[sn-mathphys-num]{sn-jnl}

\usepackage{graphicx}%
\usepackage{multirow}%
\usepackage{amsmath,amssymb,amsfonts}%
\usepackage{amsthm}%
\usepackage{mathrsfs}%
\usepackage[title]{appendix}%
\usepackage{xcolor}%
\usepackage{textcomp}%
\usepackage{manyfoot}%
\usepackage{booktabs}%
\usepackage{algorithm}%
\usepackage{algorithmicx}%
\usepackage{algpseudocode}%
\usepackage{listings}%
\usepackage{longtable}
\usepackage[center]{caption}
\usepackage{float}
\usepackage{booktabs}
\usepackage{array}
\usepackage{amsmath}
\usepackage{graphicx}
\usepackage{subcaption}

\theoremstyle{thmstyleone}%
\theoremstyle{thmstyletwo}%

\theoremstyle{thmstylethree}%

\begin{document}

\title[Article Title]{Multi-Task Learning for Lung sound \& Lung disease classification}

\author*[1]{\fnm{Suma} \sur{K V}}\email{sumakv@msrit.edu}

\author[1]{\fnm{Deepali} \sur{Koppad}}\email{deepali.koppad@gmail.com}

\author[1]{\fnm{Preethi} \sur{Kumar}}\email{preethihskumar@gmail.com}
\author[1]{\fnm{Neha A} \sur{Kantikar}}\email{nehakantikar@gmail.com}
\author[2]{\fnm{Surabhi} \sur{Ramesh}}\email{surabhirameshm@gmail.com}

\affil*[1]{\orgdiv{Department of ECE}, \orgname{Ramaiah Institute of Technology}, \orgaddress{\city{Bengaluru}, \postcode{560054}, \state{Karnataka}, \country{India}}}

\affil[2]{\orgname{M.V.J Medical College \& Research Hospital}, \orgaddress{\city{Bengaluru}, \postcode{560054}, \state{Karnataka}, \country{India}}}

\abstract{In recent years, advancements in deep learning techniques have considerably enhanced the efficiency and accuracy of medical diagnostics. In this work, a novel approach using multi-task learning (MTL) for the simultaneous classification of lung sounds and lung diseases is proposed. Our proposed model leverages MTL with four different deep learning models such as 2D CNN, ResNet50, MobileNet and Densenet to extract relevant features from the lung sound recordings. The ICBHI 2017 Respiratory Sound Database was employed in the current study. The MTL for MobileNet model performed better than the other models considered, with an accuracy of74\% for lung sound analysis and 91\% for lung diseases classification. Results of the experimentation demonstrate the efficacy of our approach in classifying both lung sounds and lung diseases concurrently. 

In this study,using the demographic data of the patients from the database, risk level computation for Chronic Obstructive Pulmonary Disease is also carried out. For this computation, three machine learning algorithms namely Logistic Regression, SVM and Random Forest classifierswere employed. Among these ML algorithms, the Random Forest classifier had the highest accuracy of 92\%.This work helps in considerably reducing the physician's burden of not just diagnosing the pathology but also effectively communicating to the patient about the possible causes or outcomes. 
}


\maketitle

\section{Introduction}\label{sec1}
Auscultation a common term in the medical field refers to listening to sound of heart and lungs. The significance of auscultation lies in the fact that it is non-invasive, cost-effective, with real-time assessment and involves patient interaction. Listening to lung sounds serves as a diagnostic tool and aids in disease management. It can also assess the severity of the lung problems and monitor the respiratory function. Lung sounds serve as a tool for identifying various lung disorders, such as  chronic obstructive pulmonary disease (COPD), bronchitis, asthma, pneumonia and pulmonary fibrosis. Abnormal lung sounds can indicate airway obstruction, presence of fluid in the lungs or inflammation. Presence of wheezing \cite{Nagasaka2012} - suggest airway constriction or narrowing and presence of crackles \cite{Pramono2017} may indicate fluid accumulation in the lungs. 

Research in the field to detect and diagnose lung sound and diseases has been going on for quite sometime now \cite {JONES199537}, \cite{Hans1997}, \cite{Bohadana2014} and many others in the recent years as well. The listening of the lung sounds is carried out at the diaphragm of the stethoscope on the chest wall, as it is more sensitive and chest sounds are relatively high pitched. The patient is advised to take deep breaths in and out through the mouth, and must be auscultated in comparable positions to each side of the torso alternatively.

The breath sounds have two main properties, intensity (loudness) and quality. They probably originate from the larger airways, are quiet and selectively filter out higher frequencies. This is termed as vesicular quality of breathing. There is no gap between the end of inspiration, and beginning of expiration. The loudness may be normal, increased or reduced. The breath sounds will be normal in intensity if the lung is expanding with air normally. In case there is localized airway narrowing, due to emphysema (permanent widening of the terminal air spaces mainly as a result of chronic smoking) or because of pleural membrane thickening (due to occupational disorders, empyema, pneumonia, infections) and pleural effusions (collection of fluid in the pleural space, there will be reduced breath
sounds. 

When there is consolidation in the lung, like in pneumonia, pulmonary oedema or pulmonary haemorrhage (area underneath the stethoscope is airless) the sounds are transmitted more efficiently, so they are louder and there is less filtering of higher frequencies. This is termed bronchial quality of breathing. The quality of the sound is harsh, the expiration is of more hissing character than inspiration, so there is a distinct differentiation between the two. Added sounds arise either in the lungs or pleura, and are abnormal. The most commonly encountered ones are wheezes and crackles. Pleural rub is another such sound characteristic of pleural inflammation, and is associated with pleural pain. It is of rubbing/ creaking nature. Other terminology like rales for coarse crackles, crepitations for fine crackles and rhonchi for wheezes can also be used as an alternate system of classification. 

Wheezes are heard mainly in expiration, as widespread and polyphonic musical sounds due to airway narrowing as in chronic obstructive pulmonary disease (COPD), asthma, infections and heart failure. Stridor is a fixed and monophonic sound indicates obstruction by a tumour or foreign body. It can be inspiratory, expiratory or both, and usually indicates need for urgent management. Crackles are short, explosive sounds similar to bubbling or clicking and are most likely due to sudden pressure changes in airways that were previously collapsed. If they appear at the beginning of inspiration, the likely pathology will be COPD. Localized, loud and coarse crackles indicate an area of pulmonary airway widening (bronchiectasis) or pulmonary oedema. In diffuse fibrosis the crackle is a late inspiratory finding, and is fine in character.

Depending on the properties of the recorded sound heard via the stethoscope, the differential diagnoses can be made and concluding the definite diagnosis becomes easier for a physician. Hence, there is a need for a non-invasive, fast and automated system that will identify the lung sound and also detect the lung disease. The proposed work explores various artificial intelligence based models for multi-task learning. In this learning, both the lung sound and lung diseases are identified and classified  simultaneously. The current work starts with ultilizing the available data of the lung sound. Train the multi-task learning model for both sound and diseases and then classifies them concurrently. The work also investigates the risk factor computation using the demographic data of the patients. 

Computation of risk factors for lung diseases involves analyzing various demographic, environmental, genetic, and behavioral factors that contribute to the likelihood of developing specific lung conditions. It involves: Gather relevant data sources, preprocess the data collected data to handle outliers, missing values, and inconsistencies. Identify the most relevant features associated with lung disease risk. Feature selection techniques  followed by, feature importance ranking and domain knowledge can assist in prioritize the most informative variables.Choose appropriate statistical or machine learning models for risk factor computation. Analyze the model coefficients or feature importance to identify the most significant risk factors linked with lung diseases. By systematically analyzing and quantifying the various risk factors connected to lung diseases, healthcare professionals can better understand the underlying mechanisms and develop targeted strategies for prevention, early detection, and management of these conditions.

Advances in lung sound research may facilitate the development of remote monitoring technologies for respiratory patients. By leveraging smart or intelligent stethoscopes (digital stethoscope) and artificial intelligence based algorithms, researchers can explore ways to accurately capture and analyze lung sounds from a distance, enabling more accessible and efficient healthcare delivery.

\textbf{Major contribution of research work}

\begin{itemize}
\item {Alter the existing deep learning models to perform multi-task learning.}
\item{Simultaneous classification of lung sounds and lung diseases.}
\item{Risk level prediction using demographic data.}
\end{itemize}

\textbf{Organization of paper}
Section two describes the current work done in the area of the proposed study. Section three discusses the methodologies adopted for multi-task learning and simultaneous classification of lung sounds and diseases. This section also elaborates on the methodology adopted for the risk factor computation. This section is followed by elaboration of results and discussions in Section four. Finally, section five summarises the study along with future scope

\section{Literature Survey}\label{seclitsur}
Using machine learning (ML) and deep learning (DL) models for analyzing lung sounds can be incredibly valuable in diagnosing and monitoring lung diseases \cite{19Xia2022}. ML and DL algorithms can automatically extract relevant features from lung sound recordings. Once features are extracted, ML and DL algorithms can be used to train classification models\cite{20Garcia2023} . Common algorithms include decision trees, k-nearest neighbors (k-NN), support vector machines (SVM),  random forests as well as deep learning architectures like convolutional neural networks (CNNs) and recurrent neural networks (RNNs) \cite{21Li2022}. ML and DL models can be augmented with techniques like data augmentation to increase the robustness of the models and improve their performance. Data augmentation involves generating synthetic data by applying transformations such as noise addition, time stretching, pitch shifting, or other variations to the original lung sound recordings. Once validated, the trained ML and DL models can be integrated into healthcare systems to assist clinicians in diagnosing lung diseases. When comparing these models for lung disease classification, it’s essential to consider the specific characteristics of the dataset, computational resources available, and the interpretability requirements of the application. Additionally, algorithms such as transfer learning, hyperparameter optimization, and ensemble learning can further improve the performance of these models \cite{22jimaging6120131} . The automated classification can be employed in smart devices which in turn can be utilized for teleconsulting applications \cite{23Huang2023}.

Several risk factors contribute to the development of lung diseases. Understanding these risk factors is crucial for prevention, early detection, and management of lung conditions. Develop a risk prediction model that combines multiple risk factors to estimate an individual's likelihood of developing a specific lung disease. This model can be used to stratify individuals into different risk categories and inform personalized prevention and intervention strategies.In addition, risk factors may also be identified for mortality prediction \cite{24Huapaya2018}.

Multi-task learning can optimize thelearning of multiple tasks that are related, concurrently. It uses the same input source to learn multiple output targets. MTLsometimes is also stated to as joint learning, and has a proximal link to other machine learning subfields such as transfer learning, and learning with auxiliary tasks and multi-class learning to name a few. A commonly followed practice to setup the inductive biasa mong different tasks is to design a parametrized hypothesis class which has some common parameters across the multiple tasks  \cite{25pmlr-v162-navon22a, 26Ozan2018}.

Joint training helps in improving data efficiency as well as in reducing computation costs. But, as the gradients of these different tasks may be conflicting, training a joint model for MTL can result in lowering of performanceas compared to its corresponding single-task counterparts. Ac ommonly applied method to avoid this challenge is merging per-task gradients into a joint update direction with the help of a particular heuristic \cite{27Thung2018}

 Table \ref{table1} presents a comparison of the work done by various researchers in the field of lung sound and lung diseases classification.

\begin{center}

 \begin{longtable}{| l | p{2.2cm} | p{2.2cm} | p{2.2cm} |  p{2.2cm}|}
\caption{Comparison of studies carried out on lung sound and lung diseases} \label{table1}\\
\hline Reference & Data \& Dataset  & Pre-processing & Techniques & Performance - Accuracy\\
\hline \cite{Ullad21}   & KRSD, CWLSD & Resampling, Zero padding MFCC, STFT-features & ANN, k-NN, Decision Tree, SVM, Random  Forest & STFT+MFCC-ANN - 98.61\% \\
\hline \cite {Jayalakshmy21} & ICBHI & intrinsic mode function (IMF), Gammatone cepstral coefficients (GTCC) & BiLSTM & Normal- 91\% Rhonchi-96\% Crackle-98\% Wheeze-94\% \\
\hline \cite{Park23}    & Self-collected  & BayesShrink denoising, MFCC, Mel-spectrogram and UMAP & SVM & 84\%  \\
\hline \cite{Gunasinghe2019} & Chest X-ray,  Kaggle and UCI Machine Learning Repository & -- & VGG16 &--  \\
\hline \cite{Goyal2023} & Chest X-ray &Median fltering, histogram equalization, visual, shape, texture, and intensity features & ANN, SVM, KNN, RNN, LSTM & F-RNN-LSTM, 93\% \\ 
\hline \cite{Mcdowell22} & Microbial extracellular vesicles, self collected & Taxonomic assignment &GLM, PCA & AUC values 0.93 to 0.99  \\
\hline \cite{Kim2021} &  Clinically collected & -- & CNN&85.7\%\\
\hline \cite{Srivastava2021} & Chest X-ray & MFCC, Mel-spectrogram, Chroma CENS & CNN & AUC value of 0.89 \\
\hline \cite {Alqudah2022} & ICBHI + KAUH & Augmentation & CNN, LSTM, CNN + LSTM & CNN+LSTM 99.8\% \\ 
\hline \cite {Fraiwan2021}&ICBHI &Wavelet smoothing, displacement artifact removal, and z-score normalization &CNN + BDLSTM& 98.26\%\\
\hline \cite {Brunese2022}&Self-collected &Statistical feature vector&Neural Network&F-Measure of 0.983\\
\hline \cite {Weiss2023}&X-ray&---&Developed a CNN model & Risk factor \\
\hline \cite {Sheikh2023}&X-ray, CT images&k-symbol Lerch transcendent functions model for  image enhancement algorithm&Alex Net, VGG16Net&98.60\% for x-ray and 98.80\% for CT scan\\
\hline\cite{Ahmad2020ANT}&Local medical cases&Lung Cancer Prediction Tool designed& Random forest, Decision trees& Specificity 90.47\%, sensitivity 100\%Accuracy 93.33\%\\
\hline\cite{Lee2023}&KNHANES, KOLD&C(cigarette smoking), I(infection), P(pollution), A(asthma), U unidentified)&Fisher’s exact test, Student’s t-test &Contribution of various Risk Factors towards COPD\\
\hline\cite{Ramahi2023}&Clinically collected&Multivariate statistical parameters&Linear mixed effect models&ppFVC mapped to disease duration, ATA, WLI \\
\hline\cite{Cho2021}&NHIS-HEALS&Pooled cohort equation (PCE)&Logistic regression, Neural Network&4 Prediction scores - FRS, PCE, SCORE, and QRISK3\\
\hline\cite{Lee20231}&SMC &Augmentation&DNN&AUROC of 0.781 for CI 95\%\\
\hline

\end{longtable}
\end{center}

\section{Methodology}
\begin{figure}[!h]
\begin{center}
  \includegraphics[width=1.0\textwidth]{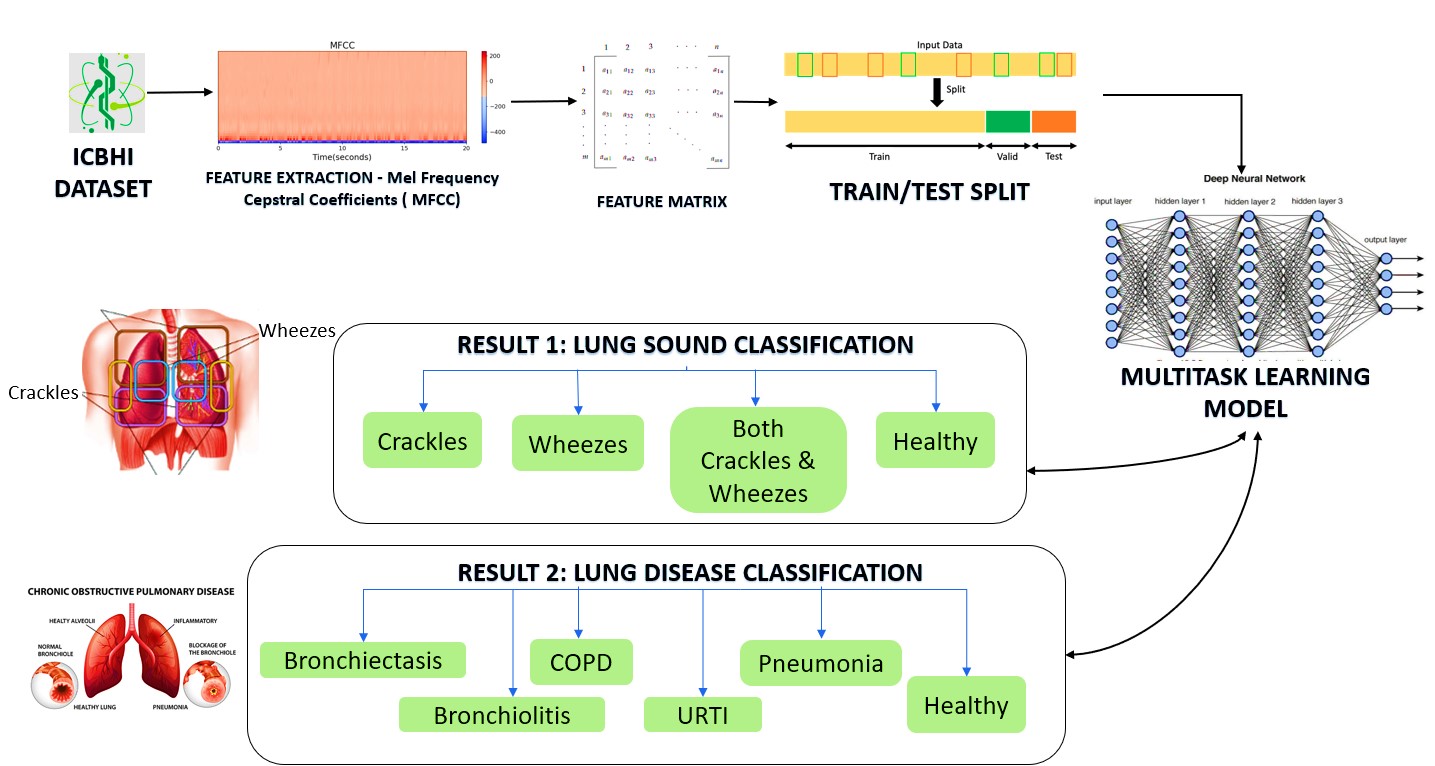}
  \caption{General Methodology of Multitask Learning}
  \label{fig3.1}
\end{center}
\end{figure}

Fig \ref{fig3.1} explains the general methodology of multitask learning used for the classification of lung sounds and diseases. The features from lung sounds present in dataset were extracted using Mel Frequency Cepstral Coefficients(MFCC), which is commonly employed for audio signal processing. The resulting feature matrix is split as training and testing data and fed to the multitask learning model which results in lung sound and lung disease classification respectively, as shown in the fig \ref{fig3.1}. Lung sound classification has four outcomes, i.e crackles, wheezes, both crackles and wheezes, and healthy. Lung disease classification results has 6 classes. Among them five are diseases, i.e Bronchiectasis, COPD, Bronchiolitis, Pneumonia,  URTI, and  and a healthy class.

The next few subsections discuss the details of each from the general methodology

\begin{enumerate}

\item \textbf{Dataset}

To implement the methodolgy explained earlier, the ICBHI 2017 Respiratory Sound Database was utilised. Lung sounds in the dataset were obtained from two to six locations in the posterior thorax using a Littman 3200 electronic stethoscope. The recordings were then transferred to a computer and saved as "wav" files. There are 920 audio recordings with annotations totaling between 10 and 90 seconds in length \cite{Rocha2019dataset}. 

\begin{table}[htbp]
\centering
\caption{Distribution of Lung sounds in the Respiratory sound database}
\label{tab:data_samples}
\begin{tabular}{|c|l|r|}
\hline
\textbf{Sl. No.} & \textbf{Lung sounds} & \textbf{Number of samples} \\ \hline
1. & Crackles & 551 \\
2. & Wheezes & 247 \\
3. & Both Crackles and Wheezes & 71 \\
4. & Healthy & 51 \\ \hline
\end{tabular}
\end{table}

There are 5.5 hours of lung sounds totaling 6898 respiratory cycles in the recordings; 886 of them include wheezes, 1864 have crackles, and 506 have both. A total of 126 individuals, including adults, seniors, and children, had their recordings made. The Patient ID, Sex, Age,  Adult BMI (kg/m2), Child Height (cm), Child Weight (kg), and Patient Number are the six columns that make up the Demographic Information File. The total number of Crackles, Wheezes, Crackles \& Wheezes, and Healthy category is shown in Table \ref{tab:data_samples}. The distribution of lung diseases is discussed in Table \ref{table_LD}.

\begin{table}[htbp]
\centering
\caption{Distribution of lung diseases in the Respiratory Sounds Database}
\label{table_LD}
\begin{tabular}{|c|l|r|}
\hline
\textbf{Sl. No.} & \textbf{Lung Diseases} & \textbf{Number of samples} \\ \hline
1. & Bronchiectasis & 16 \\
2. & Bronchiolitis & 13 \\
3. & COPD & 793 \\
4. & Pneumonia & 37 \\
5. & URTI & 23 \\
6. & Healthy & 35 \\ \hline
\end{tabular}
\end{table}

\item \textbf{Mel Frequency Cepstral Coefficients (MFCC)}

To extract cepstral coefficients as a feature from incoming lung sounds, MFCC uses multiple procedures. To amplify higher frequencies, a pre-emphasis filter is applied first. To lessen spectral leakage, the signal is then split up into brief frames and multiplied by a window function. Time frames are transformed to the frequency domain by employing Fast Fourier Transform (FFT). Subsequently, every frame's power spectrum is run through a sequence of triangle filters that are arranged according to the Mel scale. The logarithm of the filter bank energies is taken to approximate human perception of loudness. The Discrete Cosine Transform (DCT) is then applied to decorrelate the coefficients and extract the most relevant information. The lower-order DCT coefficients, known as the cepstral coefficients, are selected as the final MFCC features and stored in a feature matrix.

\item \textbf{Multitask learning}

In task learning MTL, multiple related tasks are learned simulatenously to improve the learning of a model for each task. MTL leverages the shared information among tasks to enhance the overall performance, instead of training distinct models for each task. In the context of machine learning, ”multiple tasks” refers to the scenario where we aim to learn multiple output targets using a single input source, or on the other hand, learn a single output target using multiple input sources. MTL allows for the optimization of related learning tasks simultaneously, enhancing performance and leveraging shared information across tasks.

The definition of MTL involves two factors: task relatedness, which determines how tasks are connected, and the definition of tasks, which can include supervised, unsupervised, semi-supervised, online learning, active learning, reinforcement learning, and multi-view learning tasks \cite{zhang2018overview}.

Rich Caruana \cite{caruana1998multitask} summarizes the goal of MTL as using domain information from related tasks to enhance generalization by learning tasks concurrently and utilizing a shared low-dimensional representation. The basic assumption is that learning one task can benefit the learning of other tasks, attained by jointly learning all tasks and leveraging correlated information among them to enhance individual task learning.

The following is the standard formulation for a traditional MTL algorithm \cite{thung2018brief}.
\[
W = [w^1 \, w^2 \, \dots \, w^m ] \sum_{m=1}^{M} L[X^m, y^m, w^m] + \lambda \text{Reg}(W)
\]

The task involves an \(m\)-th task, where the input matrix \(X^m\) represents the \(m\)-th task, and the output vector \(\mathbf{y}^m\) represents the corresponding \(m\)-th task output vector. The weight vector \(\mathbf{w}^m\) represents the regression parameters for the \(m\)-th task, mapping \(X^m\) to \(y^m\). The scalars \(N^m\), \(D\), and \(M\) represent the number of samples, features for each input matrix, and the number of tasks. The weight vectors in \(\{\mathbf{w}^m\}\) are concatenated to obtain \(W\), which is designed based on prior knowledge and assumptions about task relationships. The regularization parameter, \(\lambda\), controls the balance between the loss function and regularizer.

\begin{figure}[htbp]
\begin{center}
  \includegraphics[width=0.3\textwidth]{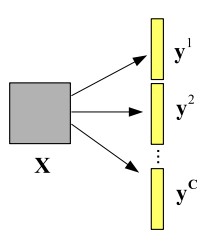}
  \caption{Single-input multi-output (SIMO)}
  \label{fig3.2}
\end{center}
\end{figure}

In our proposed methodology we use, Single-input multi-output (SIMO) type of multitask learning as shown in Fig \ref{fig3.2}. This is a method where different output targets are predicted using one input \cite{thung2018brief}. Since the input to predict lung sounds and lung diseases is the same feature matrix obtained from MFCC, the SIMO type of multitask learning is more suitable to achieve our purposed of lung sound and lung disease classification.

The benefits of multitask learning include:

\begin{itemize}
    \item Improved generalization: By learning from multiple tasks, the model can capture more diverse patterns and generalize better to unseen data.
    \item Data efficiency: Multitask learning allows the model to leverage data from multiple tasks, even if some tasks have limited data. This can help improve performance on tasks with limited training samples.
    \item Regularization: The shared representation acts as a form of regularization, preventing overfitting and improving the model's ability to generalize.
\end{itemize}

\item \textbf{Architectures of deep learning models}

In our methodology we have trained four different models, viz 2D CNN, Resnet50, MobileNet and DenseNet, to classify lung sounds and diseases using multitask learning. The concept of MTL is incorporated into these models and then trained for Lung sound and Lung diseases. The MobileNet architecture is discussed in more depth in this paper as it is contributing significantly to the achieve our objective. 

The core of MobileNet architecture is the idea of depthwise separable convolutions, which divide normal convolutions into two different processes: pointwise and depthwise convolution. Pointwise convolution combines the resultant feature maps by executing 1x1 convolutions throughout the depth dimension, whereas depthwise convolution applies a single filter independently to each input channel. With this modular method, the network's computational complexity is significantly reduced without compromising its capacity to learn significant characteristics.

The model defines a function called bottleneck Block that defines depthwise separable convolutions and pointwise separable convolutions as shown in the fig \ref{fig3.5.2}. Deathwise Separable convolution has three layers depthwise convolution 2D layer, a batch normalization and a ReLU activation layer whereas pointwise separable convolution contains 2D convolutional layer, batch normalization and ReLU activation layer.

\begin{figure}[!h]
\begin{center}
  \includegraphics[width=0.7\textwidth]{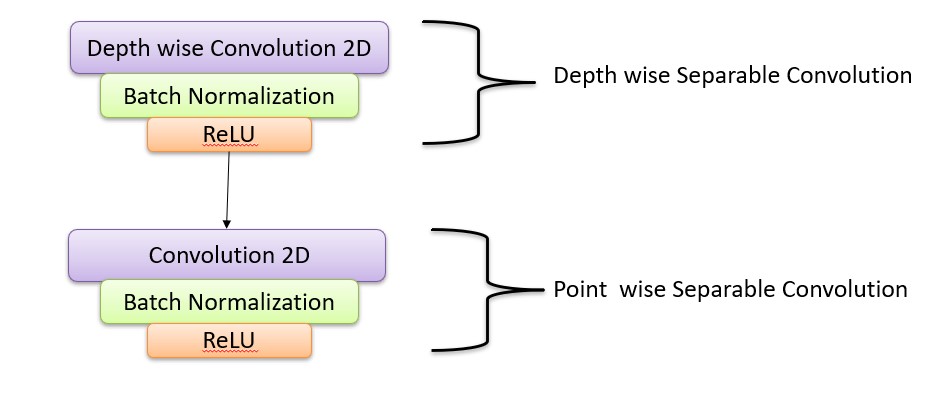}
  \caption{BottleNeck Block}
  \label{fig3.5.2}
\end{center}
\end{figure}
Fig \ref{fig3.5.1} explains that the feature matrix along with sounds labels and disease labels are given to the MobileNet model. The input passes through a  2D Convolutional layer, four bottleneck blocks with 32, 64, 128 and 256 filters respectively, a 2D Convolutional layer, a bottleneck block with 256 filters and a global average pooling layer. The outputs from global average pooling layer are concatenated. The merged features are passed through two dense layers with different activation functions ReLU and softmax, namely. The resulting outputs are classified lung sounds and lung diseases.

\begin{figure}[htbp]
\begin{center}
  \includegraphics[width=1.0\textwidth]{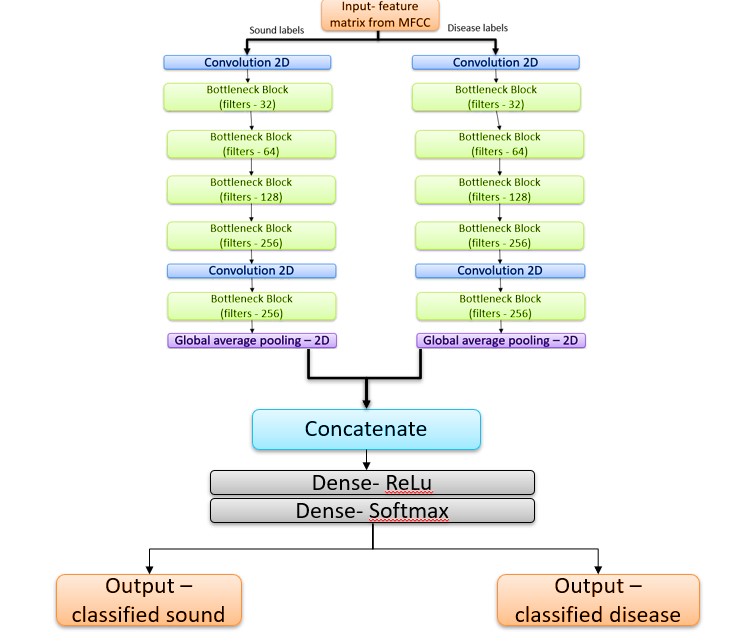}
  \caption{Architecture of MTL for MobileNet}
  \label{fig3.5.1}
\end{center}
\end{figure}

\item \textbf{Risk Levels}

Fig \ref{fig3.7} explains the proposed process used for the prediction of risk levels of the disease Chronic Obstructive Pulmonary Disease (COPD). Due to imbalance of samples of different diseases in ICBHI dataset and COPD patients dominating the dataset, we chose to predict risk levels for only COPD. Demographic data such as age, gender and BMI of patients were considered and studied to predict risk levels of COPD with the aid of machine learning algorithms.

\begin{figure}[!h]
\begin{center}
  \includegraphics[width=0.7\textwidth]{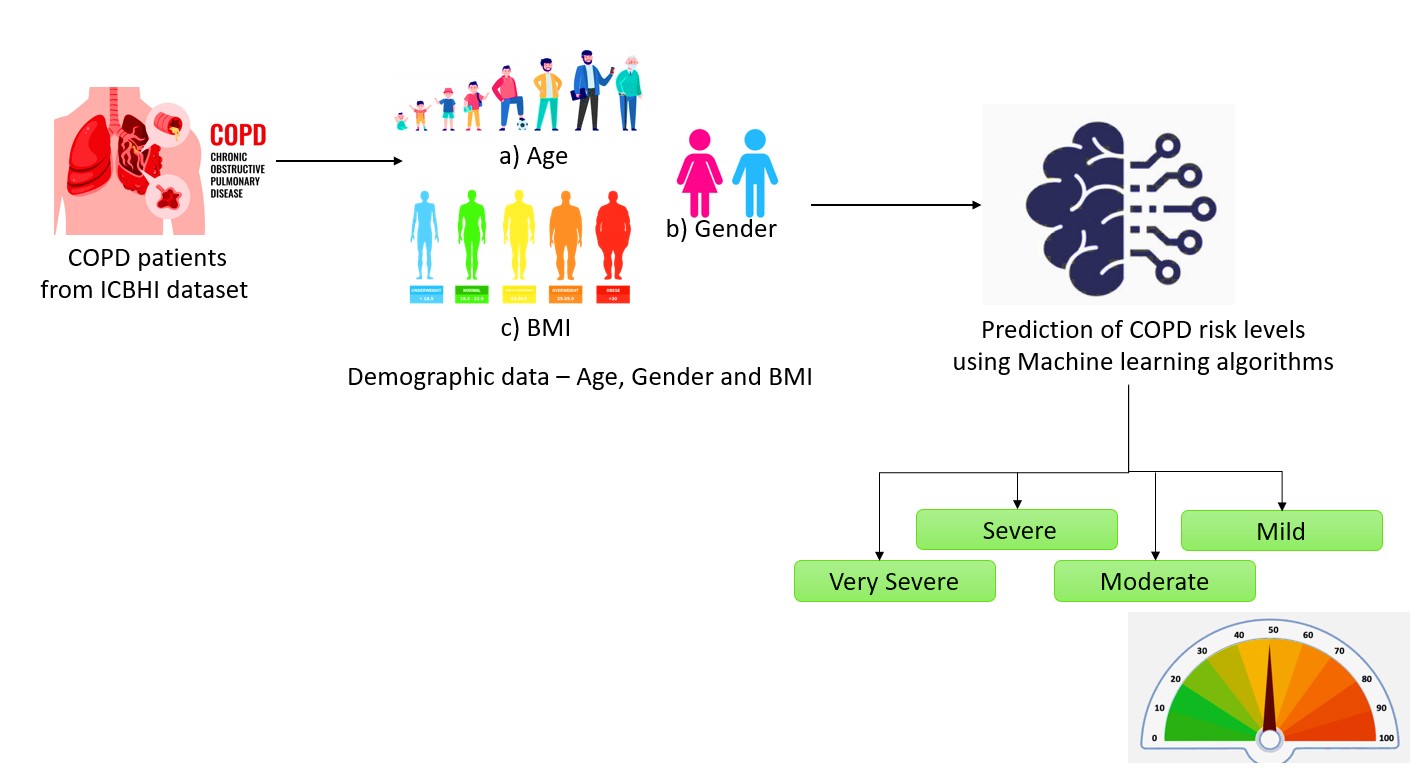}
  \caption{Prediction of COPD Risk levels}
  \label{fig3.7}
\end{center}
\end{figure}
\begin{enumerate}
\item \textbf{Risk Factors of COPD}

\begin{itemize}
\item{Age}

Aging is a significant risk factor for chronic diseases, with an increase in life expectancy leading to a rise in age-related diseases. The lung ages, resulting in dimished function and decreased capacity to react to environmental stresses.cCOPD, a condition of faster lung aging, is two to three times greater in people aged above 60 years \cite{RFMacNee2016}. Some of the mechanisms associated with aging that are present in the lungs of patients with COPD are inflammation,  oxidative stress and cell senescence. These changes may enhance the disease's activity and increase susceptibility to exacerbations. National Heart, Lung, and Blood Institute Trusted Source states that COPD is observed to be prevalent in people above 40 years who have the habit of smoking or had the habit earlier in life.

\item{Gender}

The main cause of COPD, a chronic lung illness, is tobacco use, and smoking contributes significantly to the disease's prevalence in women. Studies reveal a complex link between gender, COPD, and tobacco use, with women exhibiting a lower cumulative exposure to tobacco use than men. A genetic propensity for lung damage specific to gender in female smokers, may be present as evidenced by the faster annual progression of COPD in this population. Due to their smaller airways, women may be more exposed even if they smoke the same quantity as males. These gender disparities are also influenced by inhaling techniques, brand preference for cigarettes, and secondhand smoke.

Professor D. Sin from University of British Columbia mentions that “Women are highly symptomatic with COPD and have a marked increase in risk of lung attacks compared with male patients and as such, heightened surveillance and aggressive pharmacologic and nonpharmacologic therapies are required to enable excellent clinical outcomes in these patients,” said Don D. Sin, MD, professor of medicine at the University of British Columbia \cite{RFCarole2017}. Professor Surya Bhatt, from the University of Alabama, Birmingham mentions “The prevalence of COPD in women is fast approaching than seen in men, and airway disease may underlie some of the high COPD numbers in women that we are seeing”  \cite{RFSorheim2010}.

\item{BMI}

A meta-analysis of 30 articles found a substantial association between  COPD and BMI. The group that was overweight and obese had a lower risk of COPD, but the underweight group was at a higher risk. For each group—underweight, overweight, and obese—the pooled chances of COPD were 1.96, 0.80, and 0.86, respectively \cite{RFZhang2021}.  With high between-study variability excluded, the obese group's pooled chances of COPD were 0.77. This suggests that BMI is connected with COPD, with underweight potentially increasing the risk wile obesity and  overweight potentially reducing it.

\end{itemize}

\item \textbf{RISK LEVELS }
Based on the above studies and considering the prevalence of each risk factor on COPD, the following risk levels have been assigned as shown in Table \ref{risklevels}.

\begin{table}[h]
\centering
\caption{Risk Levels of COPD}
\label{risklevels}
\begin{tabular}{|l|l|l|l|}
\hline
\textbf{Risk Level} & \textbf{Age} & \textbf{BMI} & \textbf{Gender} \\ \hline
Class 0: Very Severe & 65+ & Underweight ($<$18.5 kg/m$^2$) & Female \\ \hline
Class 1: Severe & 65+ & Underweight ($<$18.5 kg/m$^2$) & Male \\ \hline
Class 2: Moderate & 50-64 & Healthy (18.5 < BMI < 24.9) & Male/Female \\ \hline
Class 3: Mild & 35-49 & Overweight, Obese & Male/Female \\ \hline
\end{tabular}
\end{table}

Table \ref{tab:risklevelsofcopdpatients} shows a part of assigned risk levels of COPD from the ICBHI dataset for a better understanding of our proposed methodology to predict risk levels. 

\begin{table}[h]
\centering
\caption{A snippet of Risk Levels of COPD patients from the ICBHI dataset }
\label{tab:risklevelsofcopdpatients}
\begin{tabular}{|l|l|l|l|}
\hline
\textbf{Age} & \textbf{Gender} & \textbf{BMI} & \textbf{Risk} \\ \hline
70 & 0 & 28.47 & 1 \\ \hline
73 & 0 & 21 & 1 \\ \hline
75 & 0 & 33.7 & 1 \\ \hline
84 & 0 & 33.53 & 1 \\ \hline
75 & 1 & 25.21 & 1 \\ \hline
60 & 1 & 22.86 & 2 \\ \hline
58 & 1 & 28.41 & 2 \\ \hline
77 & 1 & 23.12 & 1 \\ \hline
68 & 1 & 24.4 & 1 \\ \hline
81 & 1 & 36.76 & 1 \\ \hline
78 & 1 & 35.14 & 1 \\ \hline
65 & 1 & 29.07 & 1 \\ \hline
65 & 0 & 24.3 & 1 \\ \hline
85 & 0 & 17.1 & 0 \\ \hline
71 & 1 & 34 & 1 \\ \hline
\end{tabular}
\end{table}

\item \textbf{Prediction of Risk Levels Using Machine Learning Algorithms}
\begin{enumerate}
\item \textbf{Support Vector Machine (SVM)}

\textbf{Train-Test Data} 

The dataset was divided in the ratio of 80:20; 80\% for the training phase and 20\% for the test phase. 

\textbf{SVM Classifier}

The SVC (Support Vector Classifier) model is a popular implementation of Support Vector Machines (SVMs) for classification tasks. In our specific case, the kernel parameter is set to 'rbf', which stands for Radial Basis Function kernel. This kernel function is commonly used when the decision boundary is non-linear. The C parameter represents the regularization parameter, controlling the trade-off between achieving a low training error and a low testing error, while the gamma parameter controls the shape of the kernel function. When set to 'auto', the value of gamma is calculated as 1/n features. 

\item \textbf{Logistic Regression}

\textbf{Train-Test Data} 
The dataset was split 80:20, with 80\% designated for the training phase and 20\% for the test phase.

\textbf{Logistic Regression Classifier} 

Logistic Regression is a linear classification algorithm employed for solving binary and multiclass classification problems. We pass three parameters during instantiation: random state, multi class, and max iterations. The random state parameter sets the seed for the random number generator used by the algorithm which is set to zero in this model. The multi class parameter specifies the type of multi-class classification that should be performed, with 'multinomial' indicating that a softmax function should be used to calculate the probabilities of each class. The max iterations parameter specifies the highest number of repetitions for the solver to converge. It is set to 1000 here. Logistic Regression is a simple yet effective algorithm useful in a variety of classification tasks. Probability of the input belonging to each class is modelled followed by selection of the class with the highest probability as the predicted output

\item \textbf{Random Forest}
\textbf{Train-Test Data}
The dataset was split 80:20, with the training phase comprising 80\% and the test phase 20\%.

\textbf{Random Forest Classifier}
Random Forest Classifier is a type of ensemble learning method that uses a collection of decision trees to make predictions. It is implemented in scikit-learn, a popular Python machine learning library. The hyperparameter n\_estimators=100 specifies that the classifier should use 100 decision trees in the forest and. random\_state=42 is used to ensure that the results are reproducible. This parameter sets the random seed to a fixed value, so that the same results will be obtained every time the code is run with the same data. Overall, Random Forest Classifier is a powerful and widely-used classification algorithm that can handle complex datasets and avoid overfitting

\end{enumerate}
\end{enumerate}
\end{enumerate}

\section{Results and Discussion}\label{secresults}
Results of this study is presented in two parts namely: results for joint classification of lung sound and lung disease followed by the results for the risk factor computation. 
For the first part that is lung sound and lung disease, both classified sounds and classified diseases are equally important to get a clear idea of what a subject is undergoing. Multi task Learning has paved the way through calculation of both sound and disease labels at once. All the deep learning models are run on NVIDIA RTX A6000 version23.0.4 accelerated GPU. The results for the MobileNet network are discussed in detail here. Similar results for the other models – 2D CNN, ResNet50 and DenseNet are also obtained. The MTL approach reduced the training time of the model when compared to training these models for lung sound and lung disease separately. 

\subsection{Results}
\textbf{Multi Task Learning for Joint Classification of lung sound and lung disease}

The results obtained for MobileNet after training of data and testing on 182 samples are presented in this section. 

\begin{figure}[!h]
\begin{center}
  \includegraphics[width=1.0\textwidth]{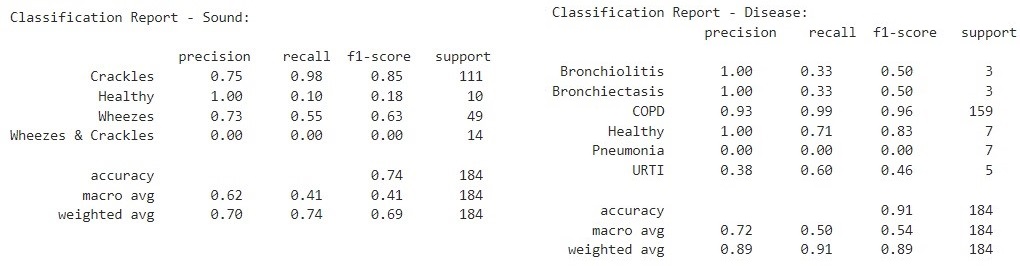}
  \caption{Classification report for lung sounds}
  \label{fig4.15}
\end{center}
\end{figure}

\textbf{Classification Report}
Fig \ref{fig4.15} depict the recall, precision,  support and  F1-score along with accuracy for both lung sound and lung disease classification.

%

The classification report depicts the recall, precision, support and F1-score for both illness and sound categorization. The MobileNet model was used to train for different batch sizes and epochs. For batch sizes of 16 and 20 epochs, 74\% classification accuracy for lung sounds was achieved. For lung illnesses, 91\% classification accuracy was obtained. MobileNet has demonstrated the highest accuracy for both sound and disease categorization among the four deep learning models.

%
%

\textbf{ROC Curve} 

\begin{figure}[!h]
\begin{center}
  \includegraphics[width=0.7\textwidth]{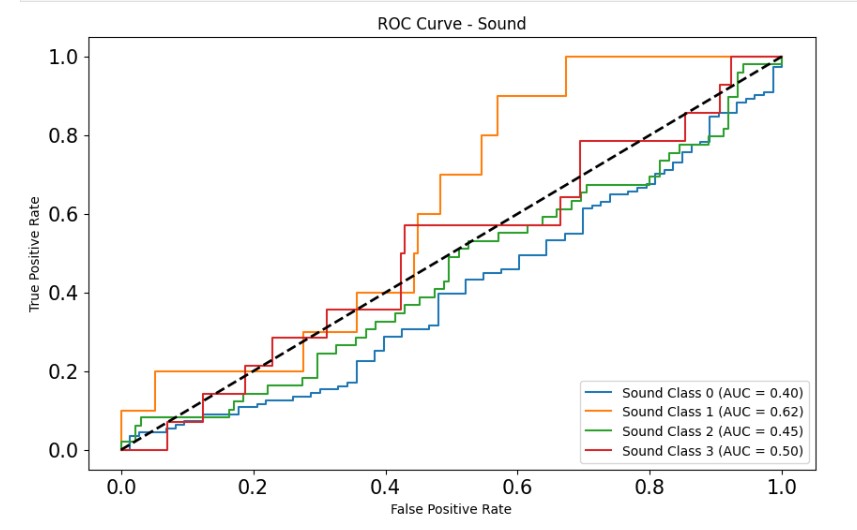}
  \caption{ROC Curve for Lung Sound Classification}
  \label{fig4.19}
\end{center}
\end{figure}

\begin{figure}[!h]
\begin{center}
  \includegraphics[width=0.7\textwidth]{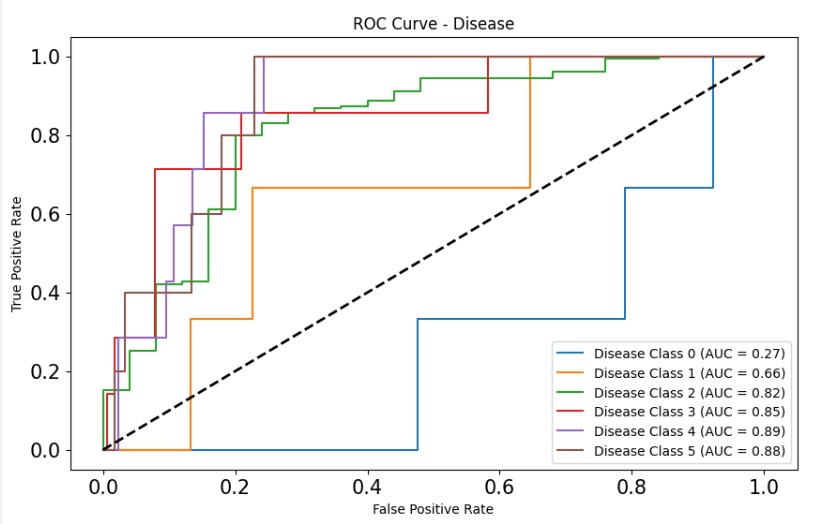}
  \caption{ROC Curve for Lung Sound Classification}
  \label{fig4.20}
\end{center}
\end{figure}

The ROC curve for MobileNet model is been shown in Fig \ref{fig4.19} and \ref{fig4.20} respectively. The area under the curve for sound in the plot of the true positive rate against the false positive rate is close to 0.5, suggesting a good model for classification. On the other hand, the performance is significantly higher for disease.

\textbf{Accuracy and Loss}

\begin{figure}[!h]
\begin{center}
  \includegraphics[width=1.0\textwidth]{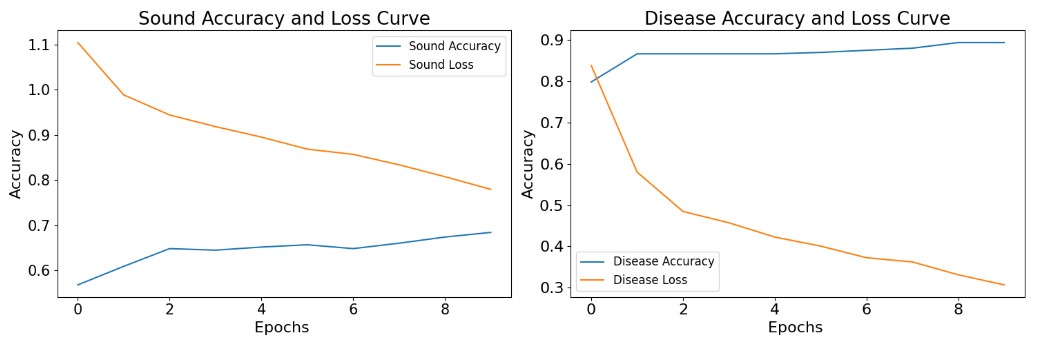}
  \caption{Accuracy and Loss Curve}
  \label{fig4.21}
\end{center}
\end{figure}

The model was trained with various epochs and batch sizes in various settings. At 20 epochs and 16 batches, an accuracy of 74\% for sound and 91\% for disease was achieved. Out of the four deep learning models, these findings indicate that MobileNet provides the highest classification accuracy. Accuracy and loss during validation on the test set are depicted in fig \ref{fig4.21}.

Similar results of classification reports, confusion matrix and ROC curves for 2D CNN, ResNet50 and DenseNet are also obtained by the authors. The consolidated results of all the models are depicted in table \ref{table-consolidatedresults}.

\begin{table}[htbp]
\centering
\caption{Performance Metrics of Different Models}
\label{table-consolidatedresults}
\begin{tabular}{|l|cc|cc|cc|cc|c|}
\hline
\multirow{2}{*}{Model} & \multicolumn{2}{c|}{Accuracy (\%)} & \multicolumn{2}{c|}{Precision} & \multicolumn{2}{c|}{Recall} & \multicolumn{2}{c|}{F1-score} & \multirow{2}{*}{Training time (s)} \\
 & LS\footnote{LS: Lung Sound} & LD\footnote{LD: Lung Disease} & LS & LD & LS & LD & LS & LD & \\ \hline
2D CNN & 66 & 89 & 0.56 & 0.81 & 0.66 & 0.89 & 0.59 & 0.84 & 24 \\
ResNet50 & 66 & 86 & 0.63 & 0.75 & 0.66 & 0.86 & 0.59 & 0.80 & 19 \\
Mobilenet & 74 & 91 & 0.70 & 0.89 & 0.74 & 0.91 & 0.69 & 0.89 & 33 \\
Densenet & 60 & 86 & 0.36 & 0.75 & 0.60 & 0.86 & 0.45 & 0.80 & 26 \\ \hline
\end{tabular}
\footnotesize
\begin{enumerate}
    \item LS: Lung Sound
    \item LD: Lung Disease
\end{enumerate}
\end{table}

The MTL MobileNet model as the best classification accuracy when compared to the other models with a lung sound (LS) accuracy of 74\% and a Lung Disease (LD) of 91\%. Though the training time for this model is slightly higher when compared to the training time of the other models.

\textbf{Results for the Risk Level Prediction}

Risk factor computation was performed using three different machine learning classifiers. Random forest, SVM and Multinomial Logistic Regression gave good results.
The risk levels were assigned namely 0, 1, 2 for very severe, severe and moderate. The results of all machine learning classifiers are as follows.
\begin{itemize}
\item The Random Forest Classifier is used to forecast the risk levels. This classifier performed well, with a 92\% accuracy rate and also the highest in comparison to other models being used for risk level prediction. 

\item The second classification model employed multinomial logistic regression, utiliz- ing logistic regression with multinomial settings for multi-class classification. This approach yielded a satisfactory accuracy of 77\%. 

\item The input variables—age, gender, and BMI—remain the same. The third machine learning method for risk level classification was the SVM model. On the testing dataset, it produced a reasonable accuracy of 61\%.
\end{itemize}

The table \ref{table-ML-RF} summarises the accuracies for the classifiers employed in the risk level computation. As it can be observed from the table, the random forest provides the  highest accuracy of 92\% in comparison to the other models. This model has a precision of 0.82, recall of -.92 and an F1-score of 0.89. 

\begin{table}[h]
\centering
\caption{Machine Learning Model Performance}
\label{table-ML-RF}
\begin{tabular}{|l|c|c|c|c|}
\hline
\textbf{ML Model} & \textbf{Accuracy (\%)} & \textbf{Precision} & \textbf{Recall} & \textbf{F1-score} \\ \hline
Random Forest & 92\% & 0.82 & 0.92 & 0.89 \\ \hline
Multinomial Logistic Regression & 77\% & 0.79 & 0.77 & 0.77 \\ \hline
SVM & 62\% & 0.38 & 0.62 & 0.47 \\ \hline
\end{tabular}
\end{table}

\subsection{Discussion}
The technique of multi-task learning was applied on four models namely, 2D CNN, Resnet 50, Mobilenet and Densenet. The performance of these models is analyzed  based on the metrics namely Accuracy, Precision, Recall and F1-score as shown in Table \ref{tab:model_performance}. It is observed that Mobilenet demonstrated the best performance with a highest accuracy of 74\% for lung sounds and 91\% for lung diseases, precision of 0.70 for lung sounds and 0.89 for lung diseases, recall of 0.74 for lung sounds and 0.91 for lung diseases and F1-score of 0.69 for lung sounds and 0.89 for lung diseases. Resnet consumed the least training time and its performance came a close second to Mobilenet. MobileNets which use depthwise convolutions have faired better in classifying sounds and diseases of lung.

\begin{table}[htbp]
\centering
\caption{Model Performance}
\label{tab:model_performance}
\begin{tabular}{|l|p{1.2cm}p{1.3cm}p{1.2cm}|p{1.2cm}p{1.3cm}p{1.2cm}|}
\hline
\multirow{2}{*}{Model} & \multicolumn{3}{c|}{Accuracy (\%)} & \multicolumn{3}{c|}{Training time (s)}  \\
 & Phase1 LS only & Phase2 LD only & Phase3 LS+LD & Phase1 LS only & Phase2 LD only & Phase3 LS+LD \\ \hline
2D CNN	&68&	90	&66 + 89	&22	&25	&24   \\
ResNet 50	&62	&86	&66 + 86	&09	&09	&19   \\
Mobilenet	&65	&92	&74 + 91	&30	&34	&33   \\
DenseNet	&60	&84	&60 + 86	&28	&23	&26  \\ \hline
\end{tabular}
\end{table}

Deep learning networks were employed to detect the four lung sounds and six types of diseases in three phases. In the first phase, the lung sounds were classified using four models, namely, 2D CNN, Resnet50, Mobilenet and Densenet. In the second phase, four lung diseases were classified using the same set of models. The input to both the phases were the lung sound files. Third phase involved the joint classification or multi-task learning of lung sounds as well as lung diseases. In this phase also, the above mentioned four models were employed. A comparison of the performance of the three phases is shown in Table \ref{tab:model_performance}. It can be observed that the accuracies obtained using MTL are comparable to the other two phases or slightly better. But there is a significant improvement in the training time. MTL results in a decrease in the time required for training as compared to the sum of training times of phase 1 and 2.

Three ML models are employed for Risk level computation as shown in Table \ref{table-ML-RF}. Among them, Random Forest provided best performance with an accuracy of 92\% , precision of 0.82 , recall of 0.92, F1 score of 0.89. Random Forest algorithm is known to be robust to overfitting, can handle nonlinear relationships well and hence seems to fair better than other models.  Having prognositc factors that put the patient at more risk of developing lung pathology, is directly proportional to prediction of the outcome.

\section{Conclusion and Future scope}

In this work we have developed a multi-task learning system which can detect different lung sounds and lung diseases simultaneously. It also provides a risk level indication by using patient demographic data. Among the four deep learning models considered, Mobilenet gave the best results for classification while Random Forest provided the best results for risk level prediction. Other classification architectures considered are 2D CNN, Resnet 50, and Densenet. Machine learning models considered for risk prediction are SVM and Multinomial logistic regresion. Performance of all the models were analyzed using Accuracy, Precision, Recall and F1-score for all the ML and DL models. Training time was an important factor for classification as multi-task learning provided an advantage compared to the individual classification conducted separately for lung sounds and lung diseases. Using this proposed analysis as aid, the physician will be able to narrow down the possible diagnosis and the stage of urgency of the lung disease.

Further, this automated analysis can be included in  intelligent stethoscopes by adopting a suitable hardware design. Development of an interactive user interface for doctor, patient and caregiver can go a long way in improving the personalized treatment. The analysis can be adopted for real-time diagnosing and monitoring of lung diseases. A multi-modal system can be developed for use in hospitals and large healthcare organizations.

%
%

\bibitem{Ullad21}Ullah, A., Khan, M., Khan, M. \& Mujahid, F. Automatic Classification of Lung Sounds Using Machine Learning Algorithms.  (2021,12)
\bibitem{Jayalakshmy21}Jayalakshmy, S. \& Sudha, G. GTCC-based BiLSTM deep-learning framework for respiratory sound classification using empirical mode decomposition.  (2021,7)
\bibitem{Park23}Park, J., Kim, K. \& Kim, J. A machine learning approach to the development and prospective evaluation of a pediatric lung sound classification model.  (2023)
\bibitem{Gunasinghe2019}Gunasinghe, A., Thirimanna, H. \& Aponso, A. Early Prediction of Lung Diseases.  (2019,3)
\bibitem{Goyal2023}Goyal, S. \& Singh, R. Early Prediction of Lung Diseases.  (2023)
\bibitem{Mcdowell22}McDowell, A., Kang, J. \& Yang, J. Machine-learning algorithms for asthma, COPD, and lung cancer risk assessment using circulating microbial extracellular vesicle data and their application to assess dietary effects.  (2022,9)
\bibitem{Kim2021}Kim, Y., Hyon, Y., Jung, S., Lee, S., Yoo, G., Chung, C. \& Ha, T. Respiratory sound classification for crackles, wheezes, and rhonchi in the clinical field using deep learning. {\em Scientific Reports}. \textbf{11} (2021,8)
\bibitem{Srivastava2021}Srivastava, A., Jain, S., Miranda, R., Patil, S., Pandya, S. \& Kotecha, K. Deep learning based respiratory sound analysis for detection of chronic obstructive pulmonary disease. {\em PeerJ Computer Science}. \textbf{7} (2021,2)
\bibitem{Alqudah2022}Alqudah, A., Qazan, S. \& Obeidat, Y. Deep learning models for detecting respiratory pathologies from raw lung auscultation sounds. . {\em Soft Computing}. \textbf{7} pp. 13405-13429  (2022,26)
\bibitem{Fraiwan2021}Fraiwan, M., Fraiwan, L., Alkhodari, M. \& Hassanin, O. Recognition of pulmonary diseases from lung sounds using convolutional neural networks and long short-term memory. {\em Journal Of Ambient Intelligence And Humanized Computing}. \textbf{13} (2021,4)
\bibitem{Brunese2022}Brunese, L., Mercaldo, F., Reginelli, A. \& Santone, A. A Neural Network-Based Method for Respiratory Sound Analysis and Lung Disease Detection. {\em Applied Sciences}. \textbf{12} pp. 3877 (2022,4)
\bibitem{Weiss2023}Weiss, J., Raghu, V. \& Al, B. Deep learning to estimate lung disease mortality from chest radiographs. {\em Nat Commun}. \textbf{14} pp. 2797 (2023)
\bibitem{Sheikh2023}Al-Sheikh, Mona \& Al, D. Multi-class deep learning architecture for classifying lung diseases from chest X-Ray and CT images. {\em Scientific Reports}. \textbf{13} (2023,11)
\bibitem{Pramono2017}Pramono, R., Bowyer, S. \& Rodriguez-Villegas, E. Automatic adventitious respiratory sound analysis: A systematic review. {\em PLoS One}. \textbf{12} (2017,5)
\bibitem{Nagasaka2012}Y, N. Lung sounds in bronchial asthma. {\em Allergology International : Official Journal Of The Japanese Society Of Allergology}. \textbf{12} pp. 353-363 (2012,3)
\bibitem{Hans1997}Hans, P., Steve S, K. \& George R, W. Respiratory Sounds: Advances Beyond the Stethoscope. {\em American Journal Of Respiratory And Critical Care Medicine}. \textbf{156} pp. 974-987 (1997)
\bibitem{Bohadana2014}Bohadana, A., Izbicki, G. \& Kraman, S. Fundamentals of Lung Auscultation. {\em The New England Journal Of Medicine}. \textbf{370} pp. 744-751 (2014,2)
\bibitem{Cho2021}Cho, S., Kim, S., Kang, S., Lee, K., Choi, D., Kang, S., Park, S., Kim, T., Yoon, C., Youn, T. \& Chae, I. Pre-existing and machine learning-based models for cardiovascular risk prediction. {\em Scientific Reports}. \textbf{11} (2021,4)
\bibitem{JONES199537}Jones, A. A Brief Overview of the Analysis of Lung Sounds. {\em Physiotherapy}. \textbf{81}, 37-42 (1995), https://www.sciencedirect.com/science/article/pii/S0031940605670344
\bibitem{Ahmad2020ANT}Ahmad, A. \& Mayya, A. A new tool to predict lung cancer based on risk factors. {\em Heliyon}. \textbf{6} (2020), https://api.semanticscholar.org/CorpusID:212552054
\bibitem{Lee2023}Lee, J., Kim, S., Kim, Y., Lee, S., Lee, J. \& Oh, Y. COPD Risk Factor Profiles in General Population and Referred Patients: Potential Etiotypes. {\em International Journal Of Chronic Obstructive Pulmonary Disease}. \textbf{Volume 18} pp. 2509-2520 (2023,11)
\bibitem{Ramahi2023}Ramahi, A., Lescoat, A., Roofeh, D., Nagaraja, V., Namas, R., Huang, S., Varga, J., O'Dwyer, D., Wang, B., Flaherty, K., Kazerooni, E. \& Khanna, D. Risk factors for lung function decline in systemic sclerosis-associated interstitial lung disease in a large single-center cohort. {\em Rheumatology (Oxford, England)}. \textbf{62} (2023,11)
\bibitem{Lee20231}Lee, Y. \& Cha, I. Multimodal deep learning of fundus abnormalities and traditional risk factors for cardiovascular risk prediction. {\em Npj Digital. Medicine}. \textbf{14} (2023,6)
\bibitem{19Xia2022}Xia, T., Han, J. \& Mascolo, C. Exploring machine learning for audio-based respiratory condition screening: A concise review of databases, methods, and open issues. {\em Experimental Biology And Medicine}. \textbf{247} pp. 153537022211154 (2022,8)
\bibitem{20Garcia2023}Garcia-Mendez, J., Lal, A., Herasevich, S., Tekin, A., Pinevich, Y., Lipatov, K., Wang, H., Qamar, S., Ayala, I., Khapov, I., Gerberi, D., Diedrich, D., Pickering, B. \& Herasevich, V. Machine Learning for Automated Classification of Abnormal Lung Sounds Obtained from Public Databases: A Systematic Review. {\em Bioengineering}. \textbf{2023} pp. 1155 (2023,10)
\bibitem{21Li2022}Li, Y., Wu, X., Yang, P., Jiang, G. \& Luo, Y. Machine Learning for Lung Cancer Diagnosis, Treatment, and Prognosis. . {\em Genomics, Proteomics & Bioinformatics}. \textbf{20} pp. 850-866 (2022,5)
\bibitem{22jimaging6120131}Kieu, S., Bade, A., Hijazi, M. \& Kolivand, H. A Survey of Deep Learning for Lung Disease Detection on Medical Images: State-of-the-Art, Taxonomy, Issues and Future Directions. {\em Journal Of Imaging}. \textbf{6} (2020), https://www.mdpi.com/2313-433X/6/12/131
\bibitem{23Huang2023}Huang, D., Huang, J., Qiao, K., Zhong, N., Lu, H. \& Wang, W. Deep learning-based lung sound analysis for intelligent stethoscope. {\em Military Medical Research}. \textbf{10} pp. 44 (2023,9)
\bibitem{24Huapaya2018}Huapaya, J., Wilfong, E., Harden, C., Brower, R. \& Danoff, S. Risk factors for mortality and mortality rates in interstitial lung disease patients in the intensive care unit. {\em European Respiratory Review}. \textbf{27} pp. 180061 (2018,12)
\bibitem{25pmlr-v162-navon22a}Navon, A., Shamsian, A., Achituve, I., Maron, H., Kawaguchi, K., Chechik, G. \& Fetaya, E. Multi-Task Learning as a Bargaining Game. {\em Proceedings Of The 39th International Conference On Machine Learning}. \textbf{162} pp. 16428-16446 (2022,7,17), https://proceedings.mlr.press/v162/navon22a.html
\bibitem{26Ozan2018}Sener, O. \& Koltun, V. Multi-Task Learning as Multi-Objective Optimization. {\em CoRR}. \textbf{abs/1810.04650} (2018), http://arxiv.org/abs/1810.04650
\bibitem{27Thung2018}Thung, C. A brief review on multi-task learning. {\em Multimed Tools Appl}. \textbf{77} pp. 29705-29725 (2018,8)
\bibitem{Rocha2019dataset}Rocha, B., Filos, D., Mendes, L., Serbes, G., Ulukaya, S., Kahya, Y., Jakovljevic, N., Turukalo, T., Vogiatzis, I., Perantoni, E., Kaimakamis, E., Natsiavas, P., Oliveira, A., Jácome, C., Marques, A., Maglaveras, N., Pedro Paiva, R., Chouvarda, I. \& Carvalho, P. An open access database for the evaluation of respiratory sound classification algorithms.. {\em Physiol Meas}. \textbf{40} (2019,3)
\bibitem{zhang2018overview}Zhang, Y. \& Yang, Q. An overview of multi-task learning. {\em National Science Review}. \textbf{5}, 30-43 (2018,1), https://doi.org/10.1093/nsr/nwx105
\bibitem{caruana1998multitask}Caruana, R. Multitask Learning. {\em Learning To Learn}. pp. 95-133 (1998)
\bibitem{thung2018brief}Thung, K. \& Wee, C. A brief review on multi-task learning. {\em Multimedia Tools And Applications}. \textbf{77} pp. 29705-29725 (2018), https://doi.org/10.1007/s11042-018-6463-x
\bibitem{RFMacNee2016}MacNee, W. Is Chronic Obstructive Pulmonary Disease an Accelerated Aging Disease?. {\em Annals Of The American Thoracic Society}. \textbf{13} pp. S429-S437 (2016,12)
\bibitem{RFCarole2017}Chrvala, C. COPD: What's Gender Got To Do With It?.  (2017,11)
\bibitem{RFSorheim2010}Sørheim, I., Johannessen, A., Gulsvik, A., Bakke, P., Silverman, E. \& DeMeo, D. Gender differences in COPD: are women more susceptible to smoking effects than men? . {\em Thorax}. (2010,6)
\bibitem{RFZhang2021}Zhang, X., Chen, H., Gu, K., Chen, J. \& Jiang, X. Association of Body Mass Index with Risk of Chronic Obstructive Pulmonary Disease: A Systematic Review and Meta-Analysis. {\em COPD}. \textbf{18} pp. 1-18 (2021,2)

\begin{thebibliography}{99}
\bibitem{Nagasaka2012}Y, N. Lung sounds in bronchial asthma. {\em Allergology International : Official Journal Of The Japanese Society Of Allergology}. \textbf{12} pp. 353-363 (2012,3)
\bibitem{Pramono2017}Pramono, R., Bowyer, S. \& Rodriguez-Villegas, E. Automatic adventitious respiratory sound analysis: A systematic review. {\em PLoS One}. \textbf{12} (2017,5)
\bibitem{JONES199537}Jones, A. A Brief Overview of the Analysis of Lung Sounds. {\em Physiotherapy}. \textbf{81}, 37-42 (1995), https://www.sciencedirect.com/science/article/pii/S0031940605670344
\bibitem{Hans1997}Hans, P., Steve S, K. \& George R, W. Respiratory Sounds: Advances Beyond the Stethoscope. {\em American Journal Of Respiratory And Critical Care Medicine}. \textbf{156} pp. 974-987 (1997)
\bibitem{Bohadana2014}Bohadana, A., Izbicki, G. \& Kraman, S. Fundamentals of Lung Auscultation. {\em The New England Journal Of Medicine}. \textbf{370} pp. 744-751 (2014,2)
\bibitem{19Xia2022}Xia, T., Han, J. \& Mascolo, C. Exploring machine learning for audio-based respiratory condition screening: A concise review of databases, methods, and open issues. {\em Experimental Biology And Medicine}. \textbf{247} pp. 153537022211154 (2022,8)
\bibitem{20Garcia2023}Garcia-Mendez, J., Lal, A., Herasevich, S., Tekin, A., Pinevich, Y., Lipatov, K., Wang, H., Qamar, S., Ayala, I., Khapov, I., Gerberi, D., Diedrich, D., Pickering, B. \& Herasevich, V. Machine Learning for Automated Classification of Abnormal Lung Sounds Obtained from Public Databases: A Systematic Review. {\em Bioengineering}. \textbf{2023} pp. 1155 (2023,10)
\bibitem{21Li2022}Li, Y., Wu, X., Yang, P., Jiang, G. \& Luo, Y. Machine Learning for Lung Cancer Diagnosis, Treatment, and Prognosis. . {\em Genomics, Proteomics \& Bioinformatics}. \textbf{20} pp. 850-866 (2022,5)
\bibitem{22jimaging6120131}Kieu, S., Bade, A., Hijazi, M. \& Kolivand, H. A Survey of Deep Learning for Lung Disease Detection on Medical Images: State-of-the-Art, Taxonomy, Issues and Future Directions. {\em Journal Of Imaging}. \textbf{6} (2020), https://www.mdpi.com/2313-433X/6/12/131
\bibitem{23Huang2023}Huang, D., Huang, J., Qiao, K., Zhong, N., Lu, H. \& Wang, W. Deep learning-based lung sound analysis for intelligent stethoscope. {\em Military Medical Research}. \textbf{10} pp. 44 (2023,9)
\bibitem{24Huapaya2018}Huapaya, J., Wilfong, E., Harden, C., Brower, R. \& Danoff, S. Risk factors for mortality and mortality rates in interstitial lung disease patients in the intensive care unit. {\em European Respiratory Review}. \textbf{27} pp. 180061 (2018,12)
\bibitem{25pmlr-v162-navon22a}Navon, A., Shamsian, A., Achituve, I., Maron, H., Kawaguchi, K., Chechik, G. \& Fetaya, E. Multi-Task Learning as a Bargaining Game. {\em Proceedings Of The 39th International Conference On Machine Learning}. \textbf{162} pp. 16428-16446 (2022,7,17), https://proceedings.mlr.press/v162/navon22a.html
\bibitem{26Ozan2018}Sener, O. \& Koltun, V. Multi-Task Learning as Multi-Objective Optimization. {\em CoRR}. \textbf{abs/1810.04650} (2018), http://arxiv.org/abs/1810.04650
\bibitem{27Thung2018}Thung, C. A brief review on multi-task learning. {\em Multimed Tools Appl}. \textbf{77} pp. 29705-29725 (2018,8)
\bibitem{Ullad21}Ullah, A., Khan, M., Khan, M. \& Mujahid, F. Automatic Classification of Lung Sounds Using Machine Learning Algorithms. (2021,12)
\bibitem{Jayalakshmy21}Jayalakshmy, S. \& Sudha, G. GTCC-based BiLSTM deep-learning framework for respiratory sound classification using empirical mode decomposition. (2021,7)
\bibitem{Park23}Park, J., Kim, K. \& Kim, J. A machine learning approach to the development and prospective evaluation of a pediatric lung sound classification model. (2023)
\bibitem{Gunasinghe2019}Gunasinghe, A., Thirimanna, H. \& Aponso, A. Early Prediction of Lung Diseases. (2019,3)
\bibitem{Goyal2023}Goyal, S. \& Singh, R. Early Prediction of Lung Diseases. (2023)
\bibitem{Mcdowell22}McDowell, A., Kang, J. \& Yang, J. Machine-learning algorithms for asthma, COPD, and lung cancer risk assessment using circulating microbial extracellular vesicle data and their application to assess dietary effects. (2022,9)
\bibitem{Kim2021}Kim, Y., Hyon, Y., Jung, S., Lee, S., Yoo, G., Chung, C. \& Ha, T. Respiratory sound classification for crackles, wheezes, and rhonchi in the clinical field using deep learning. {\em Scientific Reports}. \textbf{11} (2021,8)
\bibitem{Srivastava2021}Srivastava, A., Jain, S., Miranda, R., Patil, S., Pandya, S. \& Kotecha, K. Deep learning based respiratory sound analysis for detection of chronic obstructive pulmonary disease. {\em PeerJ Computer Science}. \textbf{7} (2021,2)
\bibitem{Alqudah2022}Alqudah, A., Qazan, S. \& Obeidat, Y. Deep learning models for detecting respiratory pathologies from raw lung auscultation sounds. . {\em Soft Computing}. \textbf{7} pp. 13405-13429 (2022,26)
\bibitem{Fraiwan2021}Fraiwan, M., Fraiwan, L., Alkhodari, M. \& Hassanin, O. Recognition of pulmonary diseases from lung sounds using convolutional neural networks and long short-term memory. {\em Journal Of Ambient Intelligence And Humanized Computing}. \textbf{13} (2021,4)
\bibitem{Brunese2022}Brunese, L., Mercaldo, F., Reginelli, A. \& Santone, A. A Neural Network-Based Method for Respiratory Sound Analysis and Lung Disease Detection. {\em Applied Sciences}. \textbf{12} pp. 3877 (2022,4)
\bibitem{Weiss2023}Weiss, J., Raghu, V. \& Al, B. Deep learning to estimate lung disease mortality from chest radiographs. {\em Nat Commun}. \textbf{14} pp. 2797 (2023)
\bibitem{Sheikh2023}Al-Sheikh, Mona \& Al, D. Multi-class deep learning architecture for classifying lung diseases from chest X-Ray and CT images. {\em Scientific Reports}. \textbf{13} (2023,11)
\bibitem{Ahmad2020ANT}Ahmad, A. \& Mayya, A. A new tool to predict lung cancer based on risk factors. {\em Heliyon}. \textbf{6} (2020), https://api.semanticscholar.org/CorpusID:212552054
\bibitem{Lee2023}Lee, J., Kim, S., Kim, Y., Lee, S., Lee, J. \& Oh, Y. COPD Risk Factor Profiles in General Population and Referred Patients: Potential Etiotypes. {\em International Journal Of Chronic Obstructive Pulmonary Disease}. \textbf{Volume 18} pp. 2509-2520 (2023,11)
\bibitem{Ramahi2023}Ramahi, A., Lescoat, A., Roofeh, D., Nagaraja, V., Namas, R., Huang, S., Varga, J., O'Dwyer, D., Wang, B., Flaherty, K., Kazerooni, E. \& Khanna, D. Risk factors for lung function decline in systemic sclerosis-associated interstitial lung disease in a large single-center cohort. {\em Rheumatology (Oxford, England)}. \textbf{62} (2023,11)
\bibitem{Cho2021}Cho, S., Kim, S., Kang, S., Lee, K., Choi, D., Kang, S., Park, S., Kim, T., Yoon, C., Youn, T. \& Chae, I. Pre-existing and machine learning-based models for cardiovascular risk prediction. {\em Scientific Reports}. \textbf{11} (2021,4)
\bibitem{Lee20231}Lee, Y. \& Cha, I. Multimodal deep learning of fundus abnormalities and traditional risk factors for cardiovascular risk prediction. {\em Npj Digital. Medicine}. \textbf{14} (2023,6)
\bibitem{Rocha2019dataset}Rocha, B., Filos, D., Mendes, L., Serbes, G., Ulukaya, S., Kahya, Y., Jakovljevic, N., Turukalo, T., Vogiatzis, I., Perantoni, E., Kaimakamis, E., Natsiavas, P., Oliveira, A., Jácome, C., Marques, A., Maglaveras, N., Pedro Paiva, R., Chouvarda, I. \& Carvalho, P. An open access database for the evaluation of respiratory sound classification algorithms.. {\em Physiol Meas}. \textbf{40} (2019,3)
\bibitem{zhang2018overview}Zhang, Y. \& Yang, Q. An overview of multi-task learning. {\em National Science Review}. \textbf{5}, 30-43 (2018,1), https://doi.org/10.1093/nsr/nwx105
\bibitem{caruana1998multitask}Caruana, R. Multitask Learning. {\em Learning To Learn}. pp. 95-133 (1998)
\bibitem{thung2018brief}Thung, K. \& Wee, C. A brief review on multi-task learning. {\em Multimedia Tools And Applications}. \textbf{77} pp. 29705-29725 (2018), https://doi.org/10.1007/s11042-018-6463-x
\bibitem{RFMacNee2016}MacNee, W. Is Chronic Obstructive Pulmonary Disease an Accelerated Aging Disease?. {\em Annals Of The American Thoracic Society}. \textbf{13} pp. S429-S437 (2016,12)
\bibitem{RFCarole2017}Chrvala, C. COPD: What's Gender Got To Do With It?. (2017,11)
\bibitem{RFSorheim2010}Sørheim, I., Johannessen, A., Gulsvik, A., Bakke, P., Silverman, E. \& DeMeo, D. Gender differences in COPD: are women more susceptible to smoking effects than men? . {\em Thorax}. (2010,6)
\bibitem{RFZhang2021}Zhang, X., Chen, H., Gu, K., Chen, J. \& Jiang, X. Association of Body Mass Index with Risk of Chronic Obstructive Pulmonary Disease: A Systematic Review and Meta-Analysis. {\em COPD}. \textbf{18} pp. 1-18 (2021,2)




\end{thebibliography}
%

\end{document}